\pdfoutput=1

\documentclass[11pt]{article}

\usepackage[final]{acl}

\usepackage{times}
\usepackage{latexsym}

\usepackage[T1]{fontenc}

\usepackage[utf8]{inputenc}

\usepackage{microtype}

\usepackage{inconsolata}

\usepackage{graphicx}
\usepackage{subcaption}

\usepackage{multirow}

\usepackage[capitalise]{cleveref}

\usepackage{amsfonts}

\title{Leaner Training, Lower Leakage: Revisiting Memorization in \\ LLM Fine-Tuning with LoRA}

\author{Fei Wang, Baochun Li\\
Department of Electrical and Computer Engineering \\
University of Toronto \\
  \texttt{silviafey.wang@utoronto.ca, bli@ece.toronto.edu} \\}

\begin{document}

\maketitle
\begin{abstract}
Memorization in large language models (LLMs) makes them vulnerable to data extraction attacks. While pre-training memorization has been extensively studied, fewer works have explored its impact in fine-tuning, particularly for LoRA fine-tuning, a widely adopted parameter-efficient method.

In this work, we re-examine memorization in fine-tuning and uncover a surprising divergence from prior findings across different fine-tuning strategies. Factors such as model scale and data duplication, which strongly influence memorization in pre-training and full fine-tuning, do not follow the same trend in LoRA fine-tuning. Using a more relaxed similarity-based memorization metric, we demonstrate that LoRA significantly reduces memorization risks compared to full fine-tuning, while still maintaining strong task performance.
\end{abstract}


\section{Introduction}\label{sec:intro}

Memorization in large language models (LLMs) has raised growing concerns, as studies have shown that these models can retain and expose training data~\citep{carlini2023quantifying,carlini2019secret,leybzon-kervadec-2024-learning,mireshghallah-2022-quantifying}. This susceptibility to memorization has led to the emergence of data extraction attacks, where adversaries exploit model generations to recover sensitive sequences from the training corpus~\citep{carilini21extracting,kassem2024alpaca,zhang-etal-2023-ethicist}.

With the rapid proliferation of open-source LLMs, fine-tuning has become a standard approach for researchers and practitioners to adapt these models for task-specific applications. Fine-tuning is often performed using private or proprietary data, such as personal medical records, internal company documents, or domain-specific knowledge bases. This raises a critical concern: while fine-tuning datasets are typically much smaller than pre-training datasets, it remains unclear whether LLMs fine-tuned on limited data are still vulnerable to extraction attacks.

While memorization in pre-training has been extensively studied, far fewer works have systematically examined its effects in fine-tuning. As a pioneering study,~\citet{mireshghallah-2022-empirical} explored how different fine-tuning strategies---full-model, head-only, and bottleneck adapter-based fine-tuning~\citep{houlsby2019parameter}---impact membership inference attack (MIA) recall~\citep{mireshghallah-2022-quantifying} and the exposure~\citep{carlini2019secret} of fine-tuning data. Their findings indicate that head-only fine-tuning leads to the highest memorization, whereas full-model and bottleneck adapter-based fine-tuning present lower but comparable risks. However, their study is constrained by the choice of evaluation metrics, as MIA recall tends to under-detect memorized data~\citep{schwarzschild2024rethinking}, potentially leading to an underestimation of actual memorization levels. In contrast,~\citet{zeng-etal-2024-exploring} analyzed memorization at the task level using a more memorization-sensitive metric based on plagiarism detection~\citep{lee-www23}. Their results show that summarization and dialogue tasks induce significantly higher memorization compared to classification and translation, highlighting how task characteristics influence memorization severity.

In this work, we extend previous research by re-examining memorization in full and head-only fine-tuning while introducing an analysis of LoRA (Low-Rank Adaptation)~\citep{hu2022lora} fine-tuning, a widely adopted parameter-efficient fine-tuning method. While LoRA is well known for its computational efficiency, its impact on memorization and data extraction risks remains largely unexplored. As an initial step, we use the same plagiarism-based memorization metric as in~\citet{zeng-etal-2024-exploring}, uncovering a surprising divergence from previous findings regarding the susceptibility of different fine-tuning methods to data extraction. Furthermore, we conduct a comprehensive evaluation using similarity-based metrics to compare LoRA, full, and head-only fine-tuning across model scales, data duplication levels, and hyperparameter configurations.

As the first empirical analysis of memorization in LoRA fine-tuning, our study extends prior work on full and head-only fine-tuning. Our results on the family of GPT-2 and Llama 3 models demonstrate that LoRA significantly reduces memorization risks compared to full fine-tuning, achieving near-zero plagiarism-based memorization while maintaining strong task performance and ensuring similarity scores remain below extraction thresholds. These findings highlight LoRA not only as a computationally efficient alternative to full fine-tuning but also as a potential privacy-preserving approach.

\section{Preliminaries}\label{sec:prelim}

\subsection{Memorization in Pre-Training vs. Fine-Tuning}

The concept of memorization in language models, as introduced by prior work~\citep{carlini2023quantifying,carilini21extracting}, defines a sequence $s$ as \textit{extractable} if there exists a prefix $c$ that, when used as a prompt, leads the model to generate $s$ with high probability. To quantify this memorization, they employ the framework of $k$-eidetic memorization, where a generated sequence $s$ is considered memorized if it is extractable and appears in at most $k$ instances within the training data.

To perform this data extraction, the adversary is assumed to have \textit{black-box} access to the language model, allowing them to generate text without direct visibility into the model's internal parameters. The extraction process begins by initializing the model with a single-token prompt containing a special start-of-sentence token, followed by autoregressive sampling. Tokens are then iteratively selected using \textit{top-$k$} sampling with $k = 40$, ensuring that only the most probable tokens are considered at each step.

In this work, we investigate memorization in fine-tuned language models using a similar data extraction setup. However, a generated sequence $s$ is considered memorized in the fine-tuned model if it is extractable and appears in the fine-tuning data. Importantly, variations in memorization and extractability evaluation methods can lead to significantly different interpretations of memorization risks. As we will demonstrate, even when adopting more relaxed memorization criteria, LoRA fine-tuning remains highly resistant to data extraction attacks.

\subsection{Model Fine-Tuning Methods}

Full fine-tuning simply updates all model parameters. Head-only fine-tuning is a lightweight approach that freezes all model layers except the final projection layer, also known as the language modeling head, which maps hidden states to token probabilities. LoRA strikes a balance by adding small trainable low-rank matrices to attention layers while keeping the rest of the model frozen, significantly reducing the number of trainable parameters. Unlike bottleneck adapter-based fine-tuning~\citep{houlsby2019parameter}, which adds extra layers between the attention and feed-forward layers of each transformer block~\citep{mireshghallah-2022-empirical}, LoRA directly modifies attention layers, achieving efficient fine-tuning with minimal computational overhead. This design enables faster adaptation, lower memory usage, and improved scalability, making it particularly well-suited for larger models and resource-constrained environments.

Mathematically, for a pre-trained weight matrix $W_0 \in \mathbb{R}^{d \times k}$, LoRA represents the fine-tuned weights as $W = W_0 + \Delta W = W_0 + \alpha \cdot BA$
where $B \in \mathbb{R}^{d \times r}$ and $A \in \mathbb{R}^{r \times k}$ are trainable low-rank matrices with rank $r \ll \min(d,k)$, and $\alpha$ is the scaling factor which  controls the magnitude of LoRA updates. The number of trainable parameters is reduced from $d \times k$ to $r \times (d + k)$, where $r$ is typically much smaller than both $d$ and $k$. During training, only $B$ and $A$ are updated, while $W_0$ remains frozen. Another key hyperparameter is the dropout rate~\cite{dropout}, which helps prevent overfitting by randomly zeroing elements during training, improving the model's generalization capabilities. Later, with provided empirical evidence, we analyze how the rank, scaling factor, and dropout rate in LoRA influence the mitigation of memorization in fine-tuned models.


\section{Related Work}\label{sec:related}

While memorization in pre-training has been widely studied, fewer works have explored memorization in the context of fine-tuning. Two notable studies,~\citet{mireshghallah-2022-empirical} and~\citet{zeng-etal-2024-exploring}, investigate this issue from different perspectives.

\citet{mireshghallah-2022-empirical} examined how different fine-tuning strategies impact memorization, comparing full-model, head-only, and bottleneck adapter-based fine-tuning~\citep{houlsby2019parameter}. Using membership inference attack (MIA) recall~\citep{mireshghallah-2022-quantifying} and the exposure metric~\citep{carlini2019secret}, they found a surprising result: fine-tuning only the model's head leads to the highest memorization, far more than full fine-tuning, even though it updates fewer parameters. This suggests that the common practice of freezing all but the final layer may introduce unexpected risks of data leakage. However, their definition of memorization and choice of evaluation metrics differ from more recent advancements in the literature and our focus. Specifically, they equate memorization with the recall of the membership inference attack on the training set, which has a high false negative rate~\citep{schwarzschild2024rethinking}, potentially leading to an underestimation of memorization. To address this limitation, our study analyzes model memorization across different fine-tuning methods using more representative and informative evaluation metrics. Furthermore, we extend this investigation to state-of-the-art fine-tuning techniques, particularly LoRA, which has not been explored in prior memorization studies.

On the other hand,~\citet{zeng-etal-2024-exploring} investigated memorization at the task level, analyzing how different fine-tuning tasks affect memorization. They found that summarization and dialogue induce significantly higher memorization than classification, reading comprehension, and translation. Additionally, they confirmed that increasing model size amplifies memorization in fine-tuning---a phenomenon previously observed in pre-training~\citep{carlini2023quantifying}---particularly in tasks prone to high memorization. However, the language models they examined, such as BART and T5, are no longer considered modern LLMs. While they did evaluate GPT-Neo-125M, its relatively small size limits its relevance to more recent large-scale models.


\section{A First Look at Memorization of LLM Fine-Tuning}
\label{sec:eval1}

\subsection{A Tentative Metric for Fine-Tuning Memorization}

To what extent can we confirm that an LLM has memorized its training data? Using appropriate evaluation criteria is crucial, as different metrics can yield highly divergent results in data extraction attacks. Defining memorization solely as the model reproducing exact wordings from the training data, as in~\citet{carlini2023quantifying}, may be too strict and could overlook instances of semantic memorization, where the model retains knowledge without verbatim reproduction. This limitation has been highlighted by~\citet{zeng-etal-2024-exploring}.

To address this,~\citet{zeng-etal-2024-exploring} adopt an automated plagiarism detection approach from~\citet{lee-www23}, which evaluates memorization through a two-step process. First, the system retrieves the \textit{top-$n$} most similar documents to a given query using the Elasticsearch~\cite{elasticsearch} ranking. Then, it applies the PAN 2014 competition-winning text alignment algorithm to detect and classify plagiarized text pairs within the identified candidate documents.

The detected plagiarism is classified into three types: \textit{verbatim plagiarism}, which copies words or phrases exactly; \textit{paraphrase plagiarism}, which modifies wording through synonym substitution, reordering, or back translation; and \textit{idea plagiarism}, which rephrases key points in a condensed or expanded form.

While this plagiarism-based evaluation framework has been effective in detecting memorization in standard fine-tuning settings,  the extent to which other fine-tuning methods exhibit memorization under this metric remains uncertain. To investigate this, we present preliminary results assessing plagiarism-based memorization across different fine-tuning approaches, including full fine-tuning, head-only tuning, and LoRA fine-tuning. Our evaluation follows the original setup and default hyperparameters of the plagiarism detection framework\footnote{https://github.com/Brit7777/LM-plagiarism}~\citep{lee-www23}.

\subsection{Minimal Plagiarism-Based Memorization in Parameter-Efficient Fine-Tuning}

Prior work has consistently shown that increasing model size and data duplication amplifies memorization during training~\citep{carlini2023quantifying,kandpal22deduplicating}. This raises an important question: do models fine-tuned with different strategies, particularly head-only and LoRA fine-tuning, exhibit memorization trends similar to those observed in pre-training?

To investigate the impact of model size, we utilize open-source pre-trained GPT-2 models~\citep{gpt2} of varying scales from Hugging Face: GPT-2 Small (124M parameters), GPT-2 Medium (355M parameters), GPT-2 Large (774M parameters), and GPT-2 XL (1.5B parameters). To examine the effect of data duplication, we introduce a hyperparameter $\rho$, which determines the duplication factor for each training sample in the fine-tuning dataset. A higher $\rho$ increases the frequency of repeated samples while maintaining the same total dataset size. For example, in a dataset of $1,000$ samples, setting $\rho = 2$ results in $500$ unique samples, each appearing twice.
To minimize the likelihood that the fine-tuning data was seen during pre-training, we select a dataset released after GPT-2's training period. The \texttt{Arxiver} dataset~\citep{acar_arxiver2024}, which contains $63.4k$ arXiv papers published between January 2023 and October 2023, is a suitable choice for evaluating memorization in a scientific writing task. 

We fine-tune our models for $10$ epochs with a learning rate of $2 \times 10^{-4}$, using the AdamW optimizer and a cosine learning rate scheduler. Training is conducted with a batch size of 4 per device, and the sequence length is set to 512 tokens for GPT-2. Specifically, in head-only fine-tuning, we update only the last two transformer layers along with the final language modeling head, which are responsible for high-level representations and directly influence the model's token predictions. For LoRA fine-tuning, we apply a rank of $r=16$, a scaling factor of $\alpha=16$, and a dropout of $0.05$ as a standard configuration.

\begin{figure}[ht!]
	\centering
    \begin{subfigure}{1\linewidth}
        \centering
        \includegraphics[width=\linewidth]{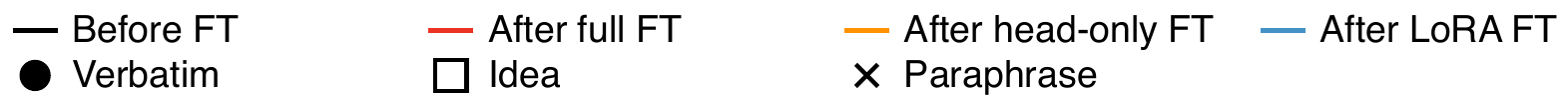}
        \label{fig:plagiarism_legend}
        \vspace{-0.4cm}
    \end{subfigure}
    \begin{subfigure}{0.5\linewidth}
        \centering
        \includegraphics[width=\linewidth]{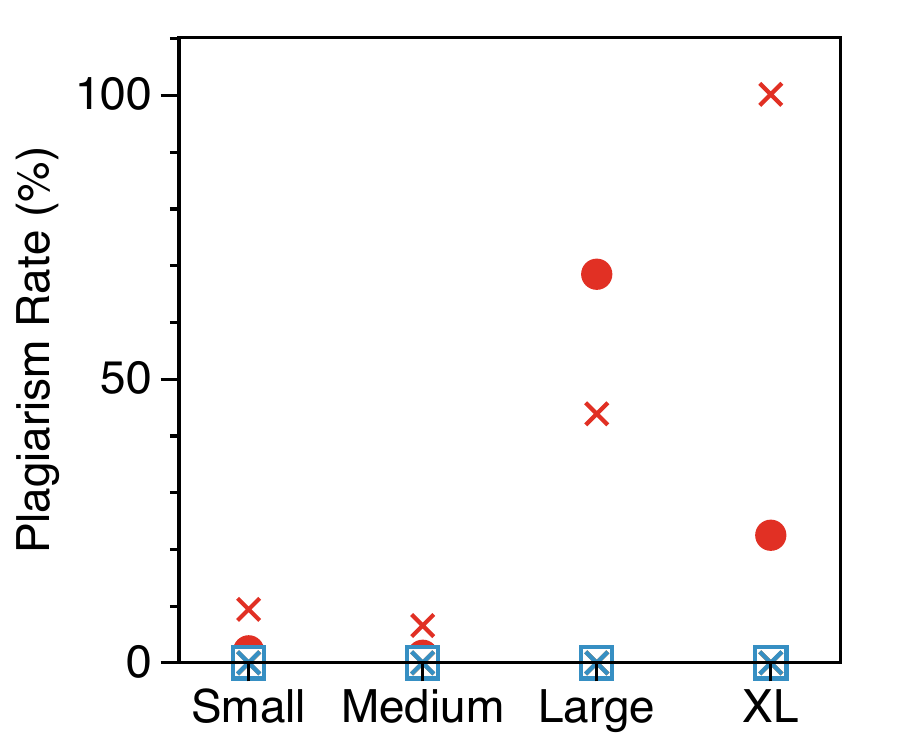}
        \caption{Model size}
        \label{fig:plagiarism_model_size}
    \end{subfigure}%
    \begin{subfigure}{0.5\linewidth}
        \centering
        \includegraphics[width=\linewidth]{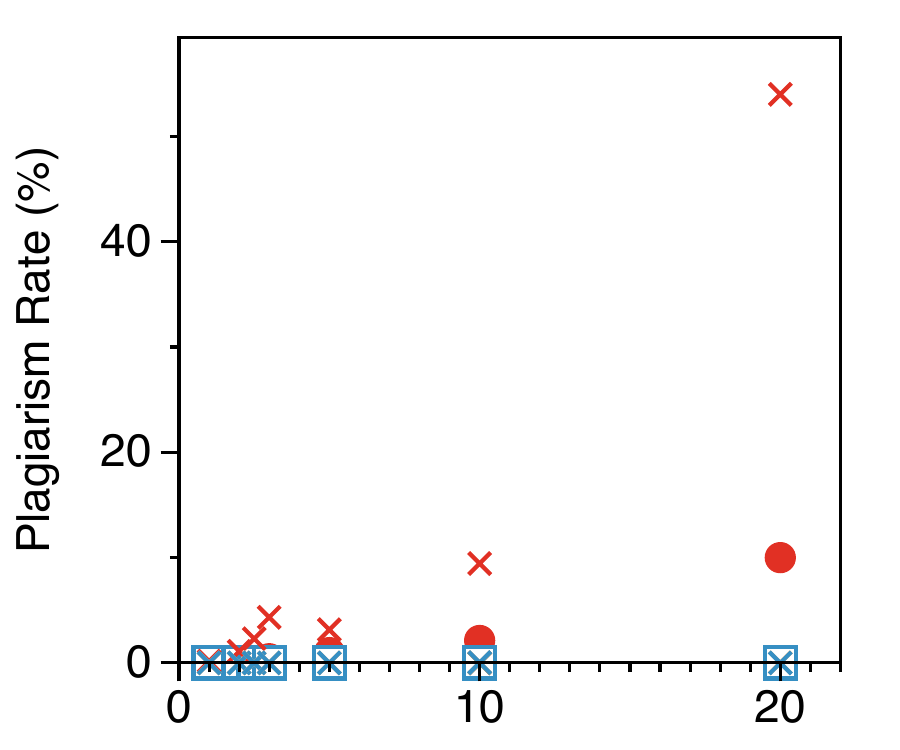}
        \caption{Data duplication level $\rho$}
        \label{fig:plagiarism_data_rep}
    \end{subfigure}
    \caption{Verbatim, idea, and paraphrase memorization across various fine-tuning methods. (a) Varying model sizes: GPT-2 Small, GPT-2 Medium, GPT-2 Large, GPT-2 XL, with fixed duplication level $\rho=10$. (b) Varying levels of data duplication ($\rho = 1, 2, 3, 5, 10, 20$) using GPT-2 Small.}
	\label{fig:plagiarism}
\end{figure}

\begin{figure}[t!]
	\centering
    \begin{subfigure}{1\linewidth}
        \centering
        \includegraphics[width=\linewidth]{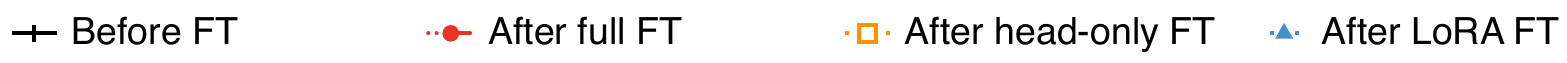}
        \label{fig:rougel_legend}
        \vspace{-0.4cm}
    \end{subfigure}
    \begin{subfigure}{0.5\linewidth}
        \centering
        \includegraphics[width=\linewidth]{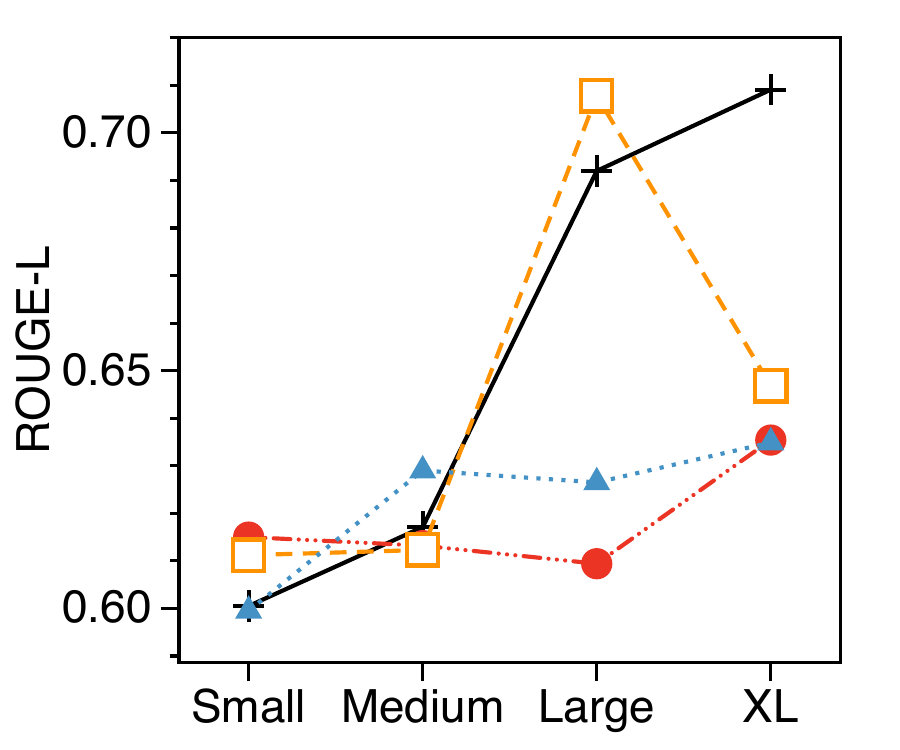}
        \caption{Model Size}
        \label{fig:rougel_model_size}
    \end{subfigure}%
    \begin{subfigure}{0.5\linewidth}
        \centering
        \includegraphics[width=\linewidth]{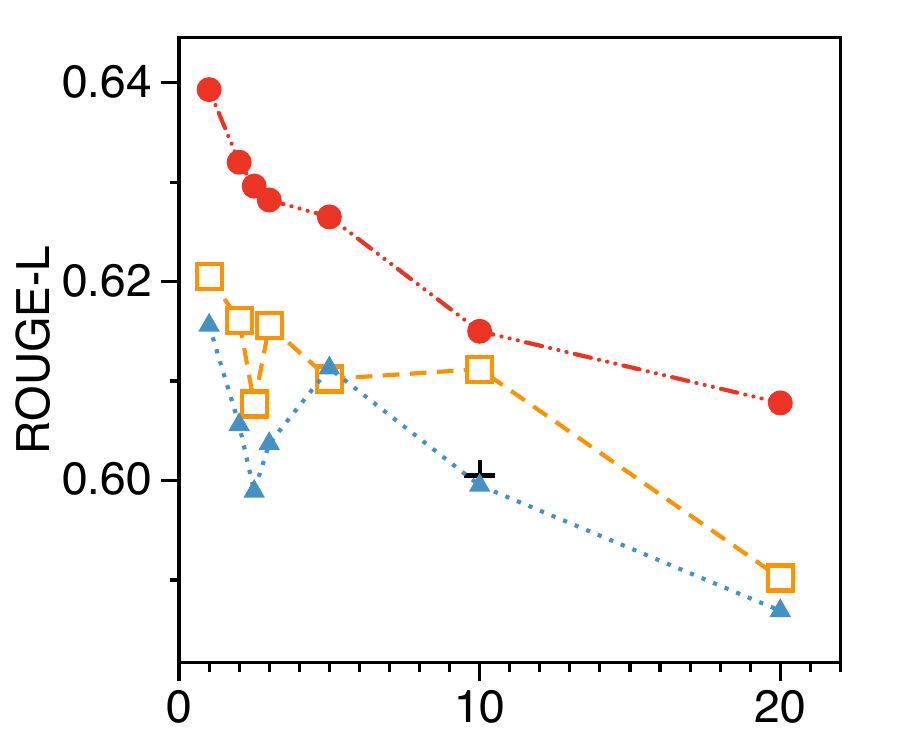}
        \caption{Data duplication level $\rho$}
        \label{fig:rougel_data_rep}
    \end{subfigure}
	\caption{ROUGE-L scores across various fine-tuning methods. (a) Varying model sizes with $\rho = 10$.
    (b) Varying duplication levels using GPT-2 Small.}
	\label{fig:rougel}
\end{figure}

We adopt the data extraction attack mentioned in~\cref{sec:prelim} to generate $1,000$ samples from each fine-tuned model and evaluate their plagiarism-based memorization. For text generation, rather than using the initial strategy of \textit{top-$k$} sampling with $k = 40$ from~\citet{carilini21extracting}, we refine token selection by setting \textit{top-$k$} to $50$ and incorporating \textit{top-$p$} sampling with $p=0.9$, ensuring that only the most probable tokens are considered while maintaining some flexibility. Additionally, we set the temperature to $0.8$ to balance randomness and diversity in the generated text.
In~\cref{fig:plagiarism}, we illustrate how different fine-tuning strategies influence plagiarism-based memorization across varying levels of data duplication and model sizes. To assess the task-specific performance of fine-tuned models, we report ROUGE-L scores as an approximate measure of model utility, shown in~\cref{fig:rougel}.


The tendency for larger models to memorize more persists in full fine-tuning, which exhibits a clear increase in verbatim and paraphrase plagiarism as model size grows. However, head-only and LoRA fine-tuning consistently maintain near-zero plagiarism rates across all model sizes. A similar trend emerges when examining the effect of data duplication. Full fine-tuning exhibits the highest plagiarism rates as $\rho$ increases, with paraphrase plagiarism surpassing 50\% at $\rho=20$. In contrast, head-only and LoRA fine-tuning remain largely unaffected, maintaining plagiarism rates close to zero even at the highest levels of redundancy. 

While full fine-tuning yields the highest ROUGE-L scores across all fine-tuning strategies, it also demonstrates the highest plagiarism rates, particularly in larger models and high-redundancy settings. Head-only and LoRA fine-tuning, on the other hand, achieve competitive ROUGE-L scores while significantly reducing plagiarism. LoRA fine-tuning, in particular, balances efficient training and reduced memorization, incurring minimal performance loss compared to full fine-tuning.

Unlike prior work, which found that head-only fine-tuning leads to the highest memorization among different fine-tuning methods for the same level of perplexity, while full fine-tuning and small adapter-based fine-tuning~\citep{houlsby2019parameter} exhibit low data leakage under MIA recall~\citep{mireshghallah-2022-quantifying} and exposure metrics~\citep{carlini2019secret}, our results reveal an entirely different trend under plagiarism-based memorization evaluation. This metric effectively captures memorization in full fine-tuning, where all model parameters are updated. However, head-only and LoRA fine-tuning consistently yield zero detected plagiarism, even as fine-tuning data duplication and model scale increase. This suggests that these parameter-efficient fine-tuning strategies, which freeze most of the model, do not exhibit direct memorization under this evaluation criterion---let alone the stricter verbatim memorization criteria originally proposed in~\citet{carilini21extracting}.


\section{Expanding the Lens: A More Permissive View of Memorization in Fine-Tuning}
\label{sec:eval2}

To further investigate memorization in fine-tuning, we extend our evaluation beyond plagiarism-based detection. A more permissive metric allows us to examine how head-only and LoRA fine-tuning strategies mitigate memorization while retaining information in a less directly extractable form. In this section, we introduce a looser memorization criterion and analyze how different fine-tuning strategies behave under this broader perspective.

\subsection{What A Looser Metric Reveals About Fine-Tuning}

To this end, we adopt \textbf{sentence similarity} to provide a more nuanced perspective of how closely generated outputs resemble training samples at a semantic level. This metric allows us to assess whether head-only and LoRA fine-tuning retain training data in a way that is not directly extractable but still recognizable through semantic similarity.

For each sequence generated from the fine-tuned model during the data extraction attack, we encode both the generated samples and the fine-tuning corpus into query embeddings and document embeddings, respectively, using the All-MPNet-Base-V2 model\footnote{https://huggingface.co/sentence-transformers/all-mpnet-base-v2}, a widely used sentence embedding model that maps text into a 768-dimensional dense vector space. We then compute cosine similarity scores between the query embeddings and document embeddings in a single matrix operation, efficiently capturing semantic similarity. The highest similarity score for each generated sequence with respect to a training sample serves as a quantitative measure of how closely the generated sentence aligns with a training instance. A cosine similarity threshold of $0.8$ is typically used to indicate strong semantic similarity. If the top sentence similarity score of a generated sequence exceeds this threshold, it is considered to closely align with a training sample from the fine-tuning dataset.

\begin{figure}[ht!]
    \centering
    \begin{subfigure}{1\linewidth}
        \centering
        \includegraphics[width=\linewidth]{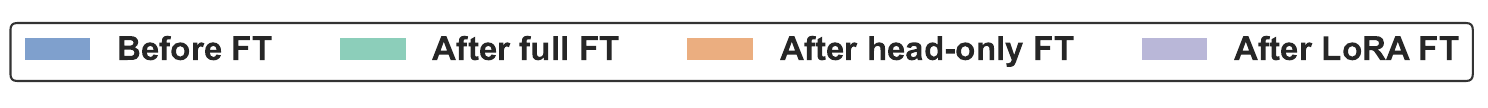}
    \end{subfigure}
    \begin{subfigure}{0.5\linewidth}
        \centering
        \includegraphics[width=\linewidth]{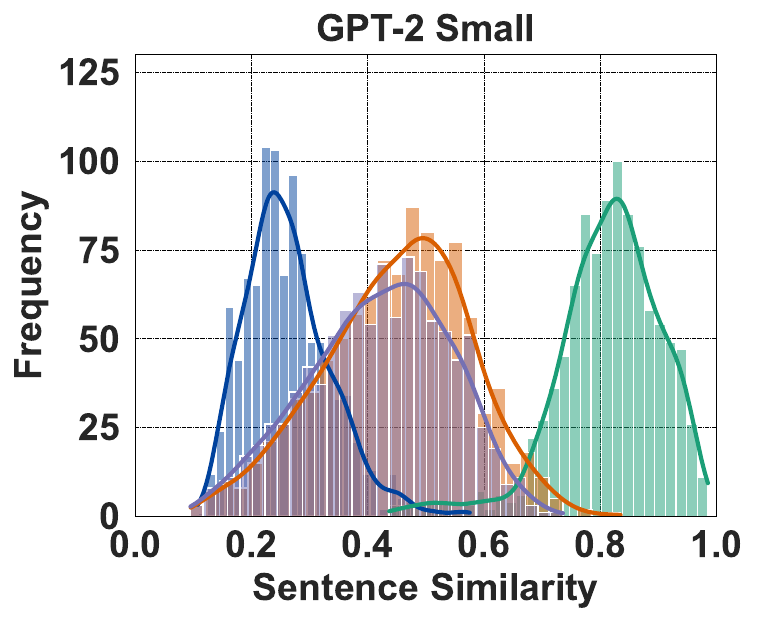}
        \label{fig:gpt2-small-arxiv-sfttrainer}
    \end{subfigure}%
    \begin{subfigure}{0.5\linewidth}
        \centering
        \includegraphics[width=\linewidth]{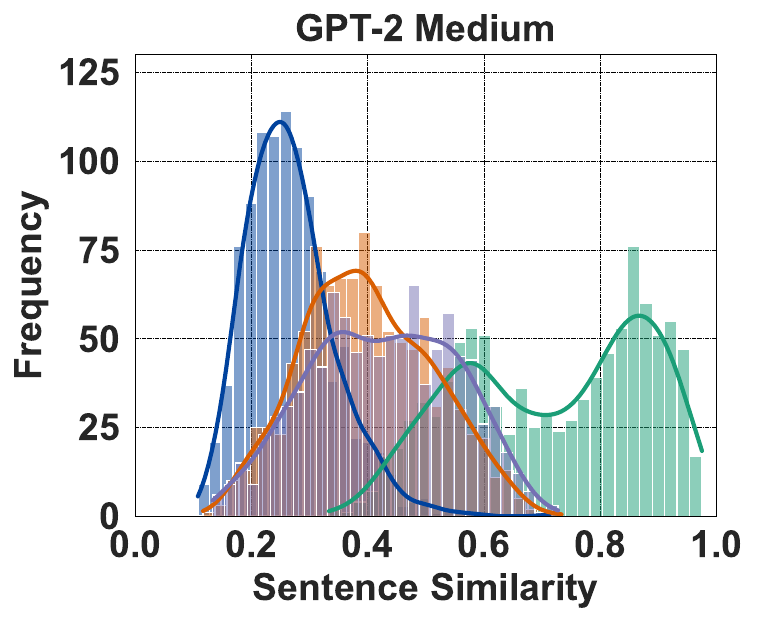}
        \label{fig:gpt2-medium-arxiv-sfttrainer}
    \end{subfigure}

    \begin{subfigure}{0.5\linewidth}
        \centering
        \includegraphics[width=\linewidth]{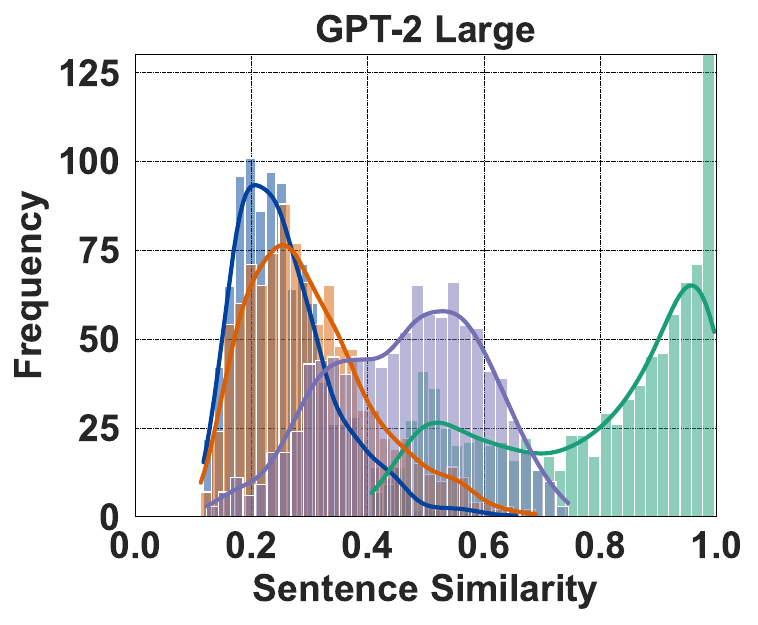}
        \label{fig:gpt2-large-arxiv-sfttrainer}
    \end{subfigure}%
    \begin{subfigure}{0.5\linewidth}
        \centering
        \includegraphics[width=\linewidth]{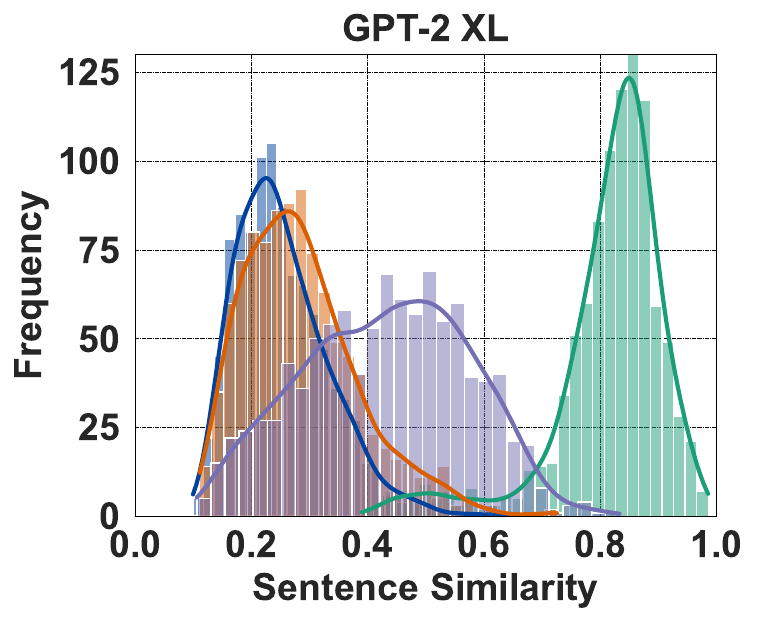}
        \label{fig:gpt2-xl-arxiv-sfttrainer}
    \end{subfigure}
    \caption{The histogram distribution of the top sentence similarity scores for $1,000$ generated samples from data extraction attacks, comparing different fine-tuning methods across various model sizes.}
    \label{fig:gpt2-arxiv-model-size-sfttrainer}
\end{figure}

\begin{figure*}[ht!]
    \centering
    \begin{subfigure}{0.6\linewidth}
        \centering
        \includegraphics[width=\linewidth]{figures/data_rep_arxiv/legend.pdf}
    \end{subfigure}
    \begin{subfigure}{0.3\linewidth}
        \centering
        \includegraphics[width=\linewidth]{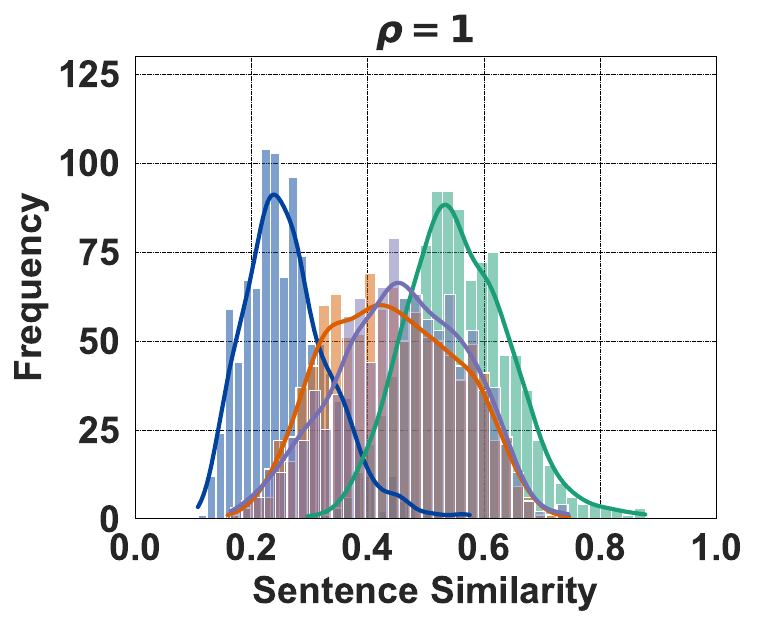}
        \label{fig:gpt2-arxiv-ss-r1.0-sfttrainer}
    \end{subfigure}%
    \begin{subfigure}{0.3\linewidth}
        \centering
        \includegraphics[width=\linewidth]{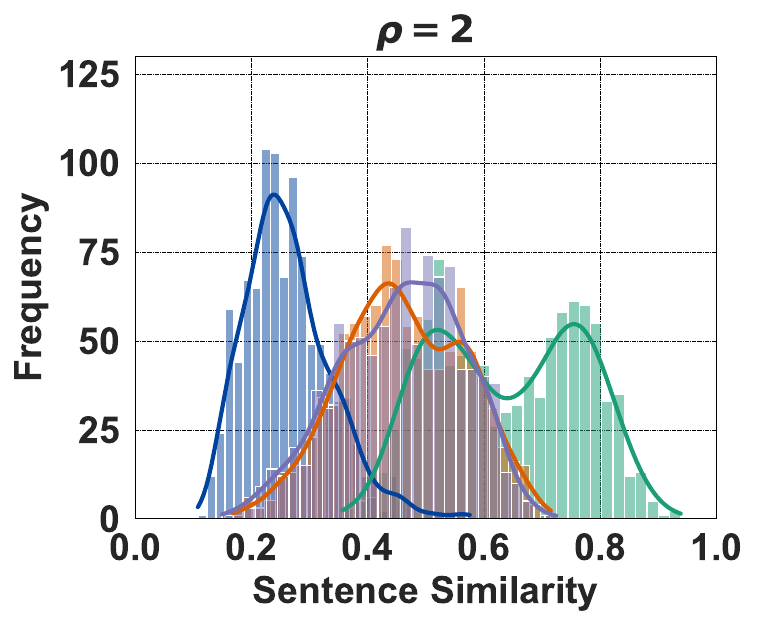}
        \label{fig:gpt2-arxiv-ss-r0.5-sfttrainer}
    \end{subfigure}%
    \begin{subfigure}{0.3\linewidth}
        \centering
        \includegraphics[width=\linewidth]{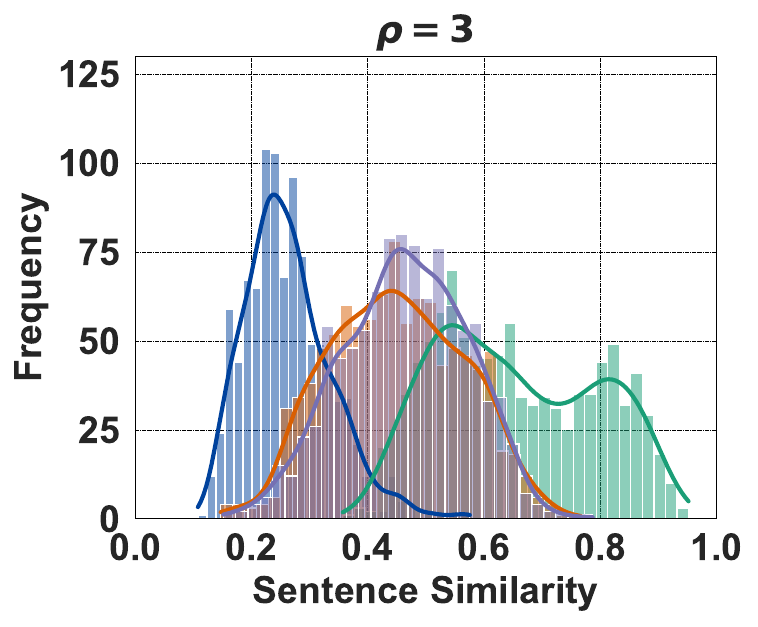}
        \label{fig:gpt2-arxiv-ss-r0.3-sfttrainer}
    \end{subfigure}%

    \begin{subfigure}{0.3\linewidth}
        \centering
        \includegraphics[width=\linewidth]{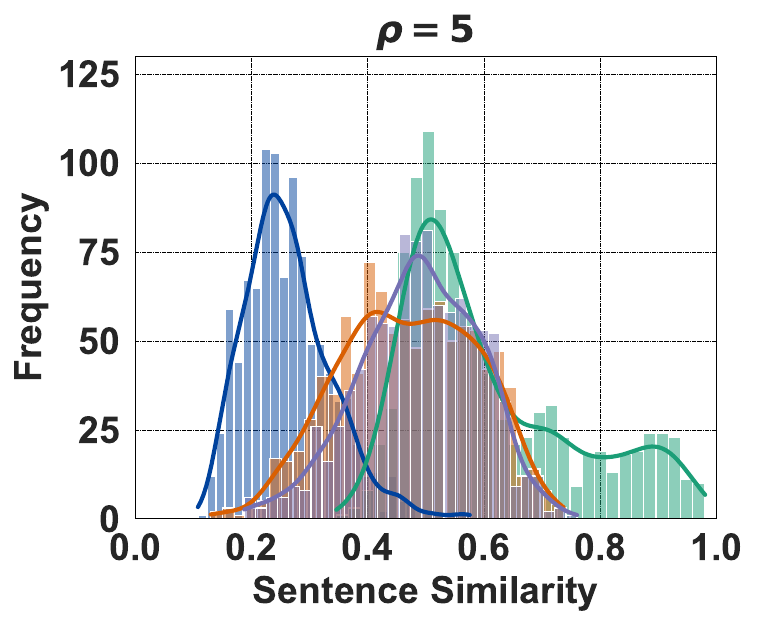}
        \label{fig:gpt2-arxiv-ss-r0.2-sfttrainer}
    \end{subfigure}%
    \begin{subfigure}{0.3\linewidth}
        \centering
        \includegraphics[width=\linewidth]{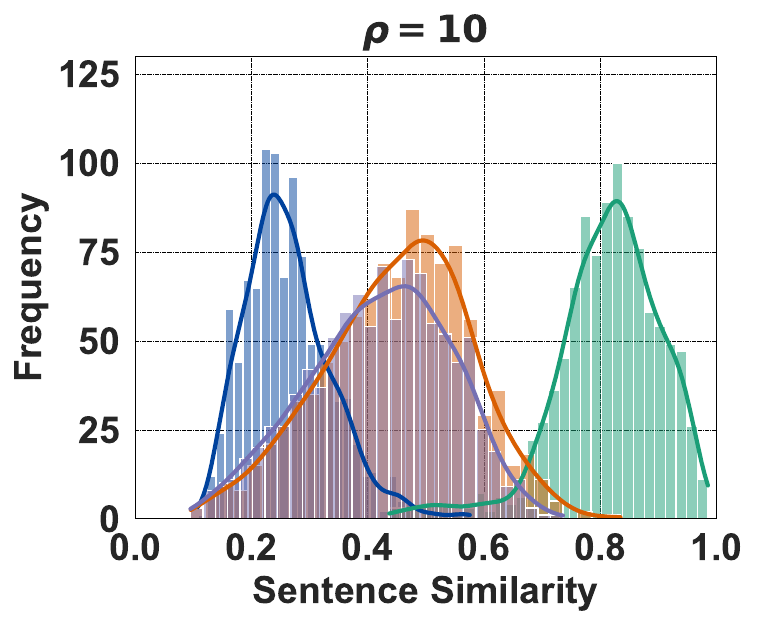}
        \label{fig:gpt2-arxiv-ss-r0.1-sfttrainer}
    \end{subfigure}%
    \begin{subfigure}{0.3\linewidth}
        \centering
        \includegraphics[width=\linewidth]{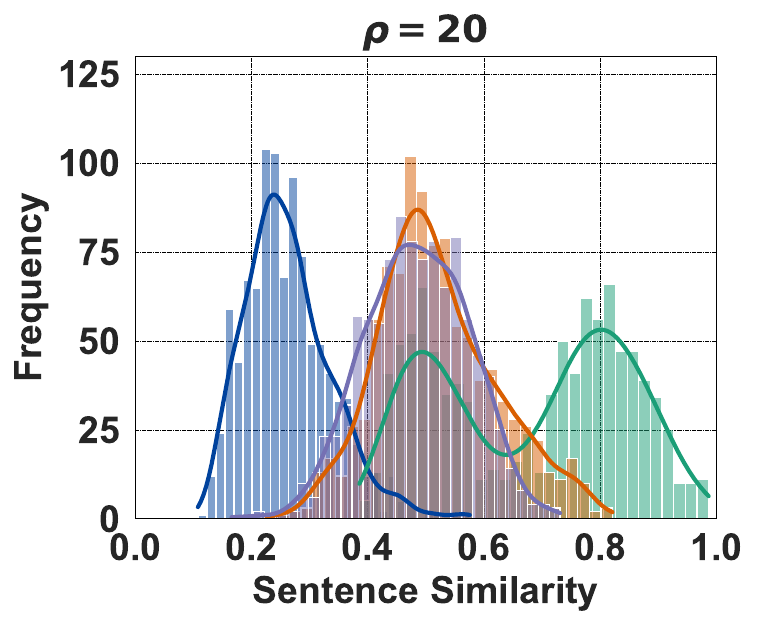}
        \label{fig:gpt2-arxiv-ss-r0.05-sfttrainer}
    \end{subfigure}
    \caption{The distribution of the top sentence similarity scores for $1,000$ generated samples from data extraction attacks, comparing different fine-tuning methods across varying levels of fine-tuning data duplication.}
    \label{fig:gpt2-arxiv-ss-rr}
\end{figure*}

As shown in~\cref{fig:gpt2-arxiv-model-size-sfttrainer}, full fine-tuning consistently shifts similarity scores rightward across all model sizes, with a pronounced concentration of samples exceeding the $0.8$ similarity threshold. This indicates a substantial increase in direct memorization compared to the pre-trained model.

Head-only fine-tuning produces outputs that remain more similar to those of the pre-trained model, especially as model size increases. This is likely because it updates only the final layers while keeping the majority of parameters frozen. As model size grows, the proportion of trainable parameters relative to the entire model decreases, further constraining the extent to which fine-tuning can reshape the model's internal representations.

LoRA fine-tuning exhibits a distinct trend compared to both full and head-only fine-tuning. While its similarity distribution shifts slightly from the pre-training baseline, it remains considerably lower than full fine-tuning, with similarity peaks concentrated below $0.6$. This suggests that LoRA enables effective adaptation while mitigating direct memorization, likely due to its parameter-efficient design, which restricts excessive retention of fine-tuning data.

For both head-only and LoRA fine-tuning, similarity distributions remain relatively stable across model sizes, with very few samples exceeding the $0.8$ threshold. This finding aligns with the plagiarism-based memorization results in~\cref{fig:plagiarism}, suggesting that samples surpassing a similarity score of $0.8$ are more likely to be flagged under plagiarism-based detection.

Similar patterns are observed in~\cref{fig:gpt2-arxiv-ss-rr}, where increasing $\rho$ in full fine-tuning shifts sentence similarity scores rightward, with more samples exceeding $0.8$. When $\rho \geq 5$, generated samples with near-perfect similarity ($\approx1.0$) begin to appear. In contrast, head-only and LoRA fine-tuning maintain stable similarity distributions with minimal deviation as $\rho$ increases. Even at the highest duplication level ($\rho=20$), the vast majority of generated samples under these methods remain below the $0.8$ similarity threshold, with distribution peaks consistently around $0.4$ to $0.5$.

\begin{figure}[ht!]
    \centering
    \begin{subfigure}{0.8\linewidth}
        \centering
        \includegraphics[width=\linewidth]{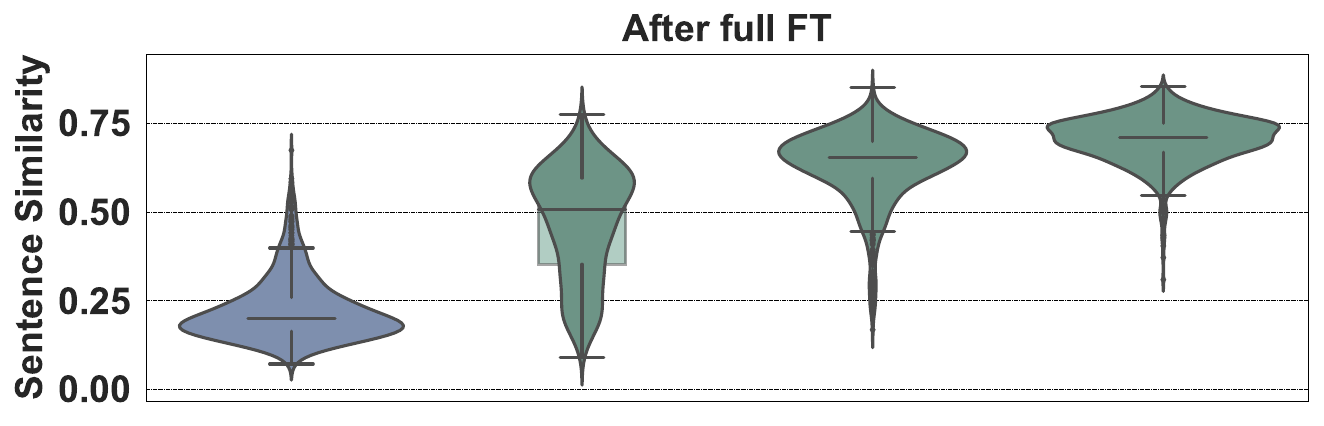}
        \label{fig:gpt2-chatdoctor-full}
    \end{subfigure}
    \begin{subfigure}{0.8\linewidth}
        \centering
        \includegraphics[width=\linewidth]{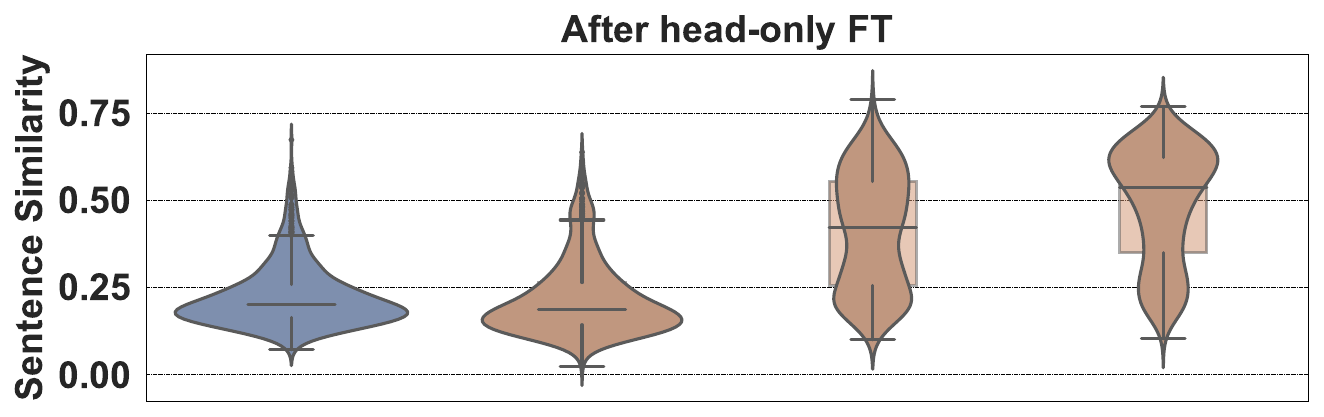}
        \label{fig:gpt2-chatdoctor-head}
    \end{subfigure}
    \begin{subfigure}{0.8\linewidth}
        \centering
        \includegraphics[width=\linewidth]{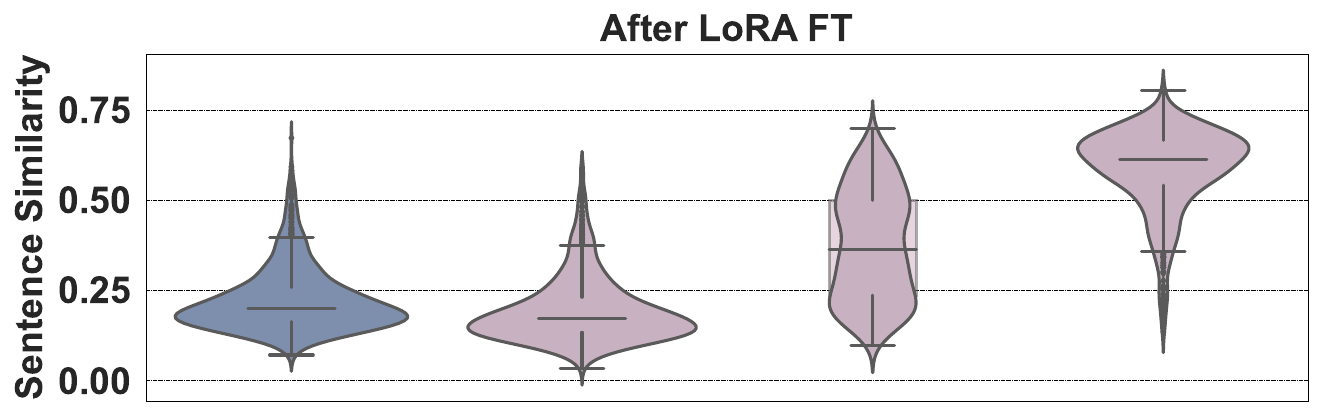}
        \label{fig:gpt2-chatdoctor-lora}
    \end{subfigure}
    \caption{Violin plots depicting the top sentence similarity scores of generated samples before and after fine-tuning GPT-2 Small on the ChatDoctor dataset. Each row represents a different fine-tuning method, while the columns correspond to training sizes of $1k$, $5k$, and $10k$ samples, respectively. The width of each violin represents the density of similarity scores at different similarity levels. The middle line marks the median.}
    \label{fig:gpt2-chatdoctor-nf}
\end{figure}

\begin{table*}[h]
    \centering
    \begin{tabular}{p{1.4cm}|p{1cm}|p{1cm}p{1cm}p{1cm}|p{1cm}p{1cm}p{1cm}|p{1cm}p{1cm}p{1cm}}
        \hline
        \multirow{2}{*}{Metrics} & Before & \multicolumn{3}{c|}{Full FT} & \multicolumn{3}{c|}{Head-only FT} & \multicolumn{3}{c}{LoRA FT}\\\cline{2-11}
        & FT & $1k$ & $5k$ & $10k$ & $1k$ & $5k$ & $10k$ & $1k$ & $5k$ & $10k$\\
        \hline
        BLEU  & $0.008$ & $0.007$ & $0.006$ & $0.006$ & $0.007$ & $0.007$ & $0.008$ & $0.007$ & $0.007$ & $0.007$ \\
        R1 & $0.208$ & $0.210$ & $0.212$ & $0.214$ & $0.210$ & $0.214$ & $0.214$ & $0.209$ & $0.212$ & $0.213$ \\
        R2  & $0.017$ & $0.017$ & $0.018$ & $0.019$ & $0.017$ & $0.017$ & $0.019$ & $0.017$ & $0.018$ & $0.018$ \\
        RL & $0.081$ & $0.080$ & $0.082$ & $0.084$ & $0.080$ & $0.082$ & $0.081$ & $0.082$ & $0.081$ & $0.082$ \\
        FRE $\uparrow$ & $58.75$ & $61.86$ & $61.71$ & $60.17$ & $63.84$ & $68.32$ & $67.70$ & $63.35$ & $64.70$ & $61.15$ \\
        SMOG $\downarrow$ & $11.32$ & $10.82$ & $10.94$ & $11.01$ & $10.51$ & $9.80$ & $9.87$ & $10.62$ & $10.39$ & $10.91$ \\
        \hline
    \end{tabular}
    \caption{Performance of models before and after fine-tuning across different fine-tuning strategies in terms of task-specific performance and linguistic complexity. R1, R2, and RL represent ROUGE-1, ROUGE-2, and ROUGE-L scores, respectively.}
    \label{tab:chatdoctor_eval}
\end{table*}

\subsection{LoRA Mitigates Memorization Even in High-Memorization Tasks}

Besides the scientific writing task using the \texttt{Arxiver} dataset, we are also interested in the high-memorization tasks explored in~\citet{zeng-etal-2024-exploring}. Following the same setup, we fine-tune GPT-2 on a dialogue task using the \texttt{ChatDoctor-HealthCareMagic} dataset~\citep{li2023chatdoctor}, which contains $112k$ user queries and doctor responses. 


\cref{fig:gpt2-chatdoctor-nf} illustrates that after full fine-tuning, sentence similarity scores increase, with the $10k$ training samples exhibiting the most concentrated distribution and the highest median value around $0.7$. In contrast, head-only fine-tuning results in a moderate increase in similarity but with greater variance, indicating a less consistent adaptation. LoRA fine-tuning follows a similar trend, maintaining a broader distribution while keeping the highest similarity score below $0.8$. This suggests that LoRA achieves a balance between adaptation and generalization, effectively preventing extractable memorization.

In addition to BLEU~\citep{bleu} and ROUGE~\citep{rouge} that quantify the similarity between the generated text and reference labels, we also measure the linguistic quality of the generated text by employing traditional readability metrics. These include Flesch Reading Ease (FRE)~\citep{Flesch1948ANR}, where higher values indicate greater readability and simpler sentence structures, and SMOG Index~\citep{mc1969smog}, which measures syntactic complexity, with higher values signifying increased difficulty.

Correspondingly, we observe in~\cref{tab:chatdoctor_eval} that full fine-tuning with larger training sizes achieves the highest content relevance for the fine-tuning task, as reflected in the highest ROUGE scores. In terms of linguistic complexity, readability improves after fine-tuning, with the FRE score increasing, especially for head-only fine-tuning. Meanwhile, the SMOG score decreases, indicating a shift toward simpler language. Overall, full fine-tuning maximizes task-specific performance but also significantly increases memorization, similar to pre-training, whereas head-only and LoRA fine-tuning maintain competitive model utility with minimal data extractability.

\subsection{Impact of LoRA Fine-Tuning Hyperparameters on Memorization}\label{sec:lora_hyper}

\begin{figure}[ht!]
	\centering
    \begin{subfigure}{0.5\linewidth}
        \centering
        \includegraphics[width=\linewidth]{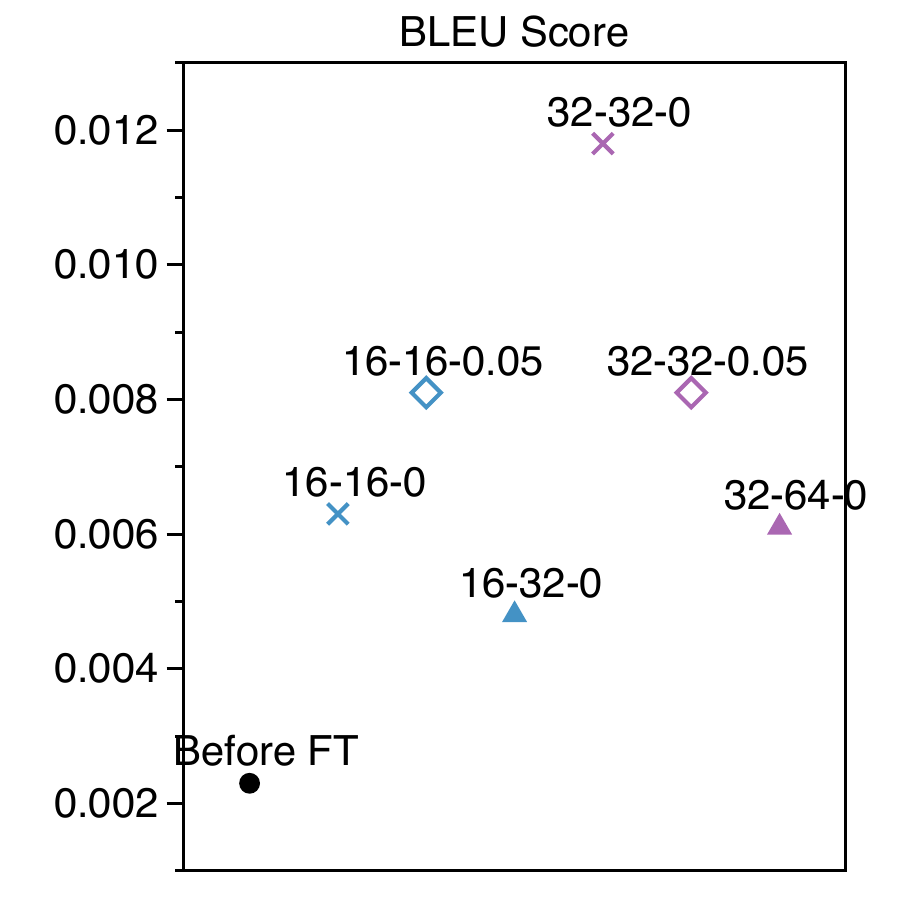}
        \caption{}
        \label{fig:llama1b_lora_bleu}
    \end{subfigure}%
    \begin{subfigure}{0.5\linewidth}
        \centering
        \includegraphics[width=\linewidth]{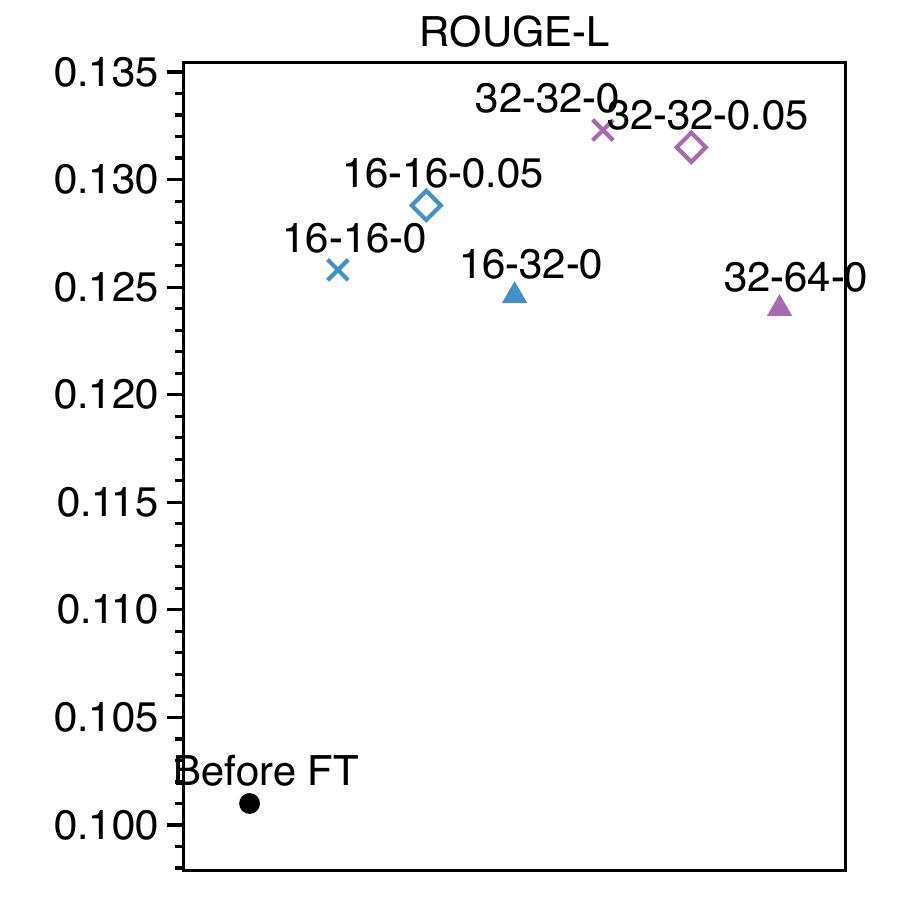}
        \caption{}
        \label{fig:llama1b_lora_rougel}
    \end{subfigure}
    \begin{subfigure}{0.8\linewidth}
        \centering
        \includegraphics[width=\linewidth]{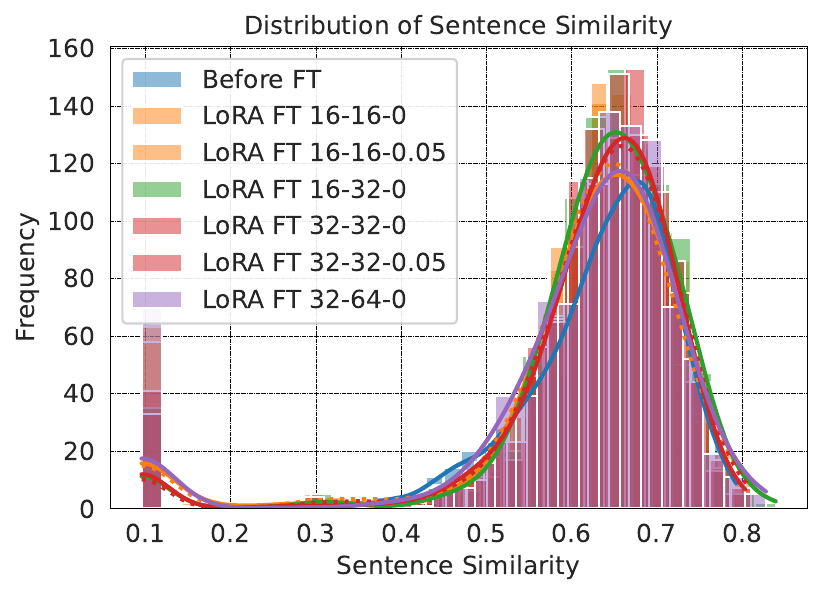}
        \caption{}
        \label{fig:llama1b_lora_sim}
    \end{subfigure}
    \caption{BLEU, ROUGE-L, and sentence similarity distribution of fine-tuned Llama 3.2 1B Instruct models under different LoRA hyperparameter settings on the ChatDoctor dataset. The label ``16-16-0.05'', for example, represents the rank $r=16$, the scaling factor $\alpha=16$, and the dropout rate $0.05$, respectively.}
	\label{fig:llama1b_lora}
\end{figure}

Having established that LoRA reduces memorization, we next explore how its hyperparameters influence both utility and extraction susceptibility. In this experiment, we tune the hyperparameters involved in LoRA fine-tuning, including the rank $r$, the scaling factor $\alpha$, and the dropout rate, to evaluate their impact on the fine-tuned model's utility and its resilience to data extraction attacks in terms of sentence similarity. We first choose to fine-tune Llama 3.2 1B model on the ChatDoctor dataset which has been split into 80\% for training and 20\% for testing. The data duplication level is set to $\rho=0$ here. \cref{fig:llama1b_lora_bleu,fig:llama1b_lora_rougel} show the model utility on the test dataset and \cref{fig:llama1b_lora_sim} shows the corresponding sentence similarity distribution of fine-tuned models under different LoRA hyperparameter settings.

All LoRA fine-tuning models show a clear improvement on model utility over the pre-trained model, and surprisingly, have lower median sentence similarity values compared to the pre-trained model without fine-tuning. Increasing the rank $r$ and the scaling factor $\alpha$ generally leads to higher scores. However, increasing the scaling factor $\alpha$ alone degrades the model utility and at the same time leads to a more concentrated sentence similarity distribution, with a few samples exceeding the $0.8$ threshold, as tagged with ``LoRA FT 16-32-0'' and ``LoRA FT 32-64-0''. This might be due to the fact that a larger $\alpha$ strengthens adaptation but may reduce generalization over the test data, and the less generalization leads to more verbatim memorization of the training data. Notably, dropout is an effective tool to further mitigate memorization without sacrificing much utility (sometimes even improving it), resulting in lower median sentence similarity values compared to configurations without dropout, consistent with its established role in improving regularization~\citep{dropout}. In overall, keeping a moderate $\alpha$ and introducing a moderate dropout rate is a good choice for LoRA fine-tuning to achieve both ideal model utility and memorization mitigation. Similar findings can also be found in GPT-2 models (see \cref{sec:more_exps}).

\begin{figure}[ht!]
    \centering
    \includegraphics[width=0.9\linewidth]{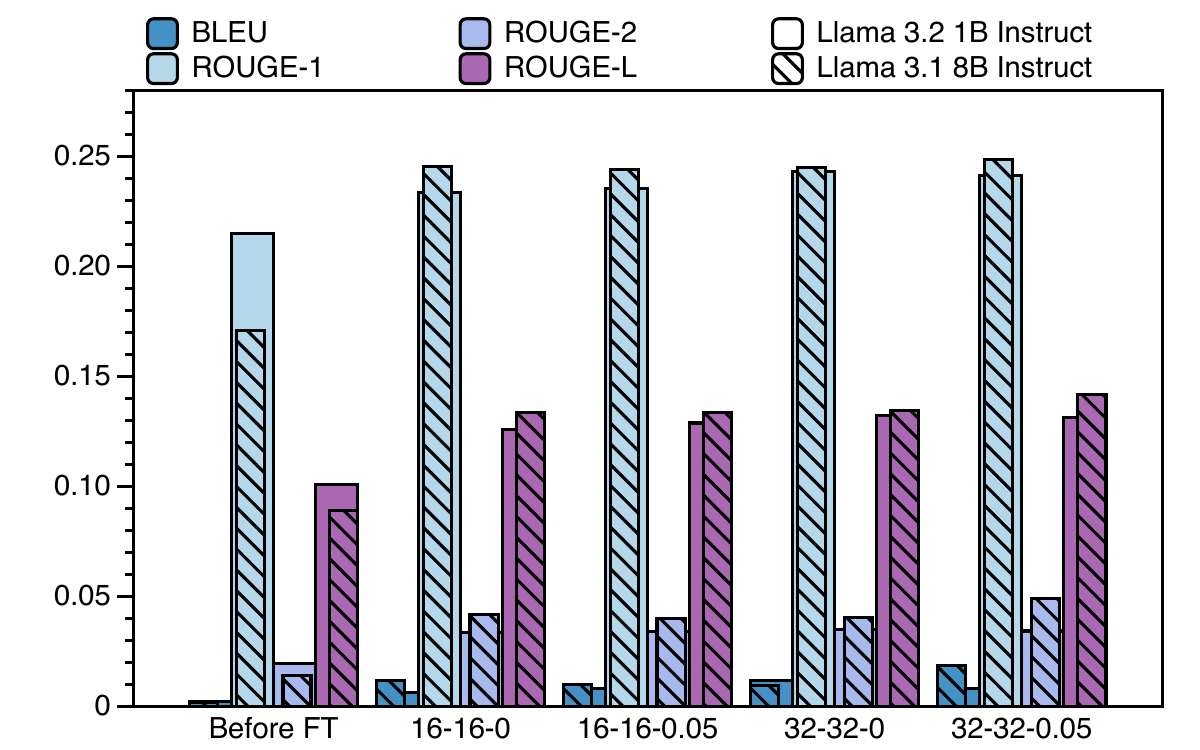}
    \caption{A comparison of BLEU, ROUGE-1, ROUGE-2, and ROUGE-L scores of Llama 3.2 1B Instruct models and Llama 3.1 8B Instruct models before and after LoRA fine-tuning.}
    \label{fig:llama_score_size}
\end{figure}

\begin{figure}[ht!]
    \centering
    \includegraphics[width=0.8\linewidth]{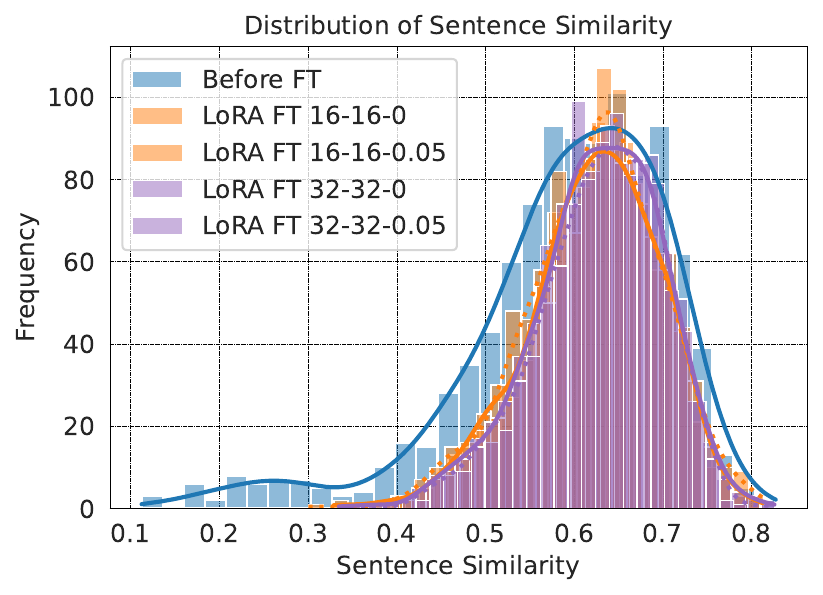}
    \caption{Sentence similarity distribution of fine-tuned Llama 3.1 8B Instruct models under different LoRA hyperparameter settings on the ChatDoctor dataset.}
    \label{fig:llama8b_lora_sim}
\end{figure}

We also conduct experiments with various LoRA hyperparameter settings on a larger Llama model, Llama 3.1 8B Instruct, and show the model utility in \cref{fig:llama_score_size} and the sentence similarity distribution in \cref{fig:llama8b_lora_sim}. The larger Llama 3.1 8B Instruct model consistently achieves higher scores than the 1B model after fine-tuning. We can also see that even with the increased model scale, the memorization mitigation effect of LoRA is still pronounced, where fewer than 5 generated samples from the fine-tuned models exceed the $0.8$ threshold. 


\section{Concluding Remarks}\label{sec:concl}

In this paper, we revisit the impact of different fine-tuning strategies on LLM memorization. Our findings show that LoRA fine-tuning, designed for computationally efficient and effective training, achieves performance comparable to full fine-tuning while significantly reducing memorization risks. Notably, it leads to near-zero plagiarism-based memorization and effectively prevents strict verbatim memorization. Across various model sizes, data duplication levels, and hyperparameter configurations, LoRA fine-tuning consistently results in lower maximum similarity scores, demonstrating its ability to mitigate memorization risks. These findings contrast with prior work, highlighting a different perspective on fine-tuning’s role in model memorization. 


\section{Limitations}\label{sec:limit}

While our study provides empirical insights into the impact of LoRA fine-tuning on model memorization, especially its key hyperparameters such as the rank, the scaling factor, and the dropout rate, a more rigorous theoretical analysis is needed to explain why LoRA fine-tuning consistently maintains memorization below the threshold of extractable memorization.

While our findings suggest that LoRA fine-tuning mitigates memorization risks, it remains unclear whether more advanced data extraction attacks, such as those proposed in~\citet{zhang-etal-2023-ethicist,kassem2024alpaca}, could enhance the extractability of fine-tuning data. Future research should investigate the interaction between LoRA fine-tuning and advanced extraction methods to determine whether further safeguards are needed to prevent memorization risks.


\bibliography{main}

\appendix



\section{Additional Experiments}\label{sec:more_exps}

The following is the additional experiment results on the impact of LoRA hyperparameters on memorization, conducted with GPT-2 model on the Arxiver dataset. \cref{tab:lora_similarity_stats} demonstrates the higher rank and scaling factors tend to improve ROUGE-L scores, with the configuration $r=32$, $\alpha=32$, dropout$=0.05$, achieving the highest ROUGE-L score of $0.5921$. The sentence similarity statistics reveal that models with a lower rank ($r=4$) and smaller scaling factors generally exhibit lower mean similarity scores, suggesting reduced memorization. However, models with higher ranks and larger scaling factors tend to have slightly higher maximum similarity values, though still well below extraction thresholds.

\begin{table*}[ht]
    \centering
    \begin{tabular}{c|c|c|c|c|c|c|c|c|c}
        \hline
        $r$ & $\alpha$ & Dropout & BLEU & ROUGE-L & Mean & Std & Min & Max & Range \\
        \hline
        4  & 16  & 0.05 & 0.4507 & 0.5881 & 0.4792 & 0.0842 & 0.1567 & 0.7046 & 0.5479 \\
        16 & 16  & 0.05 & 0.4456 & 0.5854 & 0.4641 & 0.0888 & 0.1068 & 0.7513 & 0.6445 \\
        32 & 32  & 0.05 & 0.4528 & 0.5921 & 0.4639 & 0.0795 & 0.1519 & 0.7337 & 0.5818 \\
        4  & 8   & 0.1  & 0.4408 & 0.5772 & 0.4485 & 0.0923 & 0.1833 & 0.6761 & 0.4928 \\
        8  & 16  & 0.1  & 0.4442 & 0.5816 & 0.4621 & 0.0971 & 0.1611 & 0.7194 & 0.5583 \\
        16  & 16  & 0.1  & 0.4434 & 0.5837 & 0.4660 & 0.0905 & 0.1474 & 0.7047 & 0.5573 \\
        \hline
    \end{tabular}
    \caption{BLEU and ROUGE-L scores, along with the statistical summary of sentence similarity scores, for fine-tuned GPT-2 Small models under different LoRA hyperparameter settings on the Arxiver dataset.}
    \label{tab:lora_similarity_stats}
\end{table*}

Overall, increasing rank and scaling factor enhances content alignment (higher ROUGE-L scores) but introduces a slight increase in memorization risk. Dropout helps regularize memorization, reducing extreme similarities while maintaining task performance, reinforcing its role in mitigating overfitting-induced memorization. Despite these variations, all maximum similarity scores remain below $
0.8$, supporting the conclusion that LoRA fine-tuning effectively minimizes extractable memorization compared to full fine-tuning. This suggests that LoRA’s low-rank adaptation mechanism naturally constrains memorization, even when hyperparameters are adjusted for stronger adaptation.

\section{Computational Resources}
All GPT-2 models are fine-tuned and evaluated on a server equipped with a single NVIDIA RTX A6000 GPU (48GB VRAM), 21 vCPUs, and 83GB of RAM. All Llama models are fine-tuned and evaluated on a server equipped with a single NVIDIA A100 PCIe GPU (80GB VRAM), 16 vCPUs, and 188GB of RAM.

\end{document}